\documentclass[sigconf]{acmart}

\AtBeginDocument{%
  }

\usepackage{hyperref}
\usepackage{makecell} 
\usepackage{microtype}
\usepackage{graphicx}
\usepackage{subcaption}
\usepackage{booktabs} 
\usepackage[linesnumbered,algoruled,boxed,lined]{algorithm2e}
\usepackage{subcaption}
\usepackage{balance}
\usepackage{tabularx,ragged2e}
\usepackage{multirow}

\setcopyright{acmlicensed}
\copyrightyear{2026}
\acmYear{2026}
\acmDOI{XXXXXXX.XXXXXXX}
\acmConference[KDD '26]{ACM SIGKDD Conference on Knowledge Discovery and Data Mining}{August 09--13,
  2026}{Jeju, Korea}
\acmISBN{978-1-4503-XXXX-X/2018/06}

\begin{document}

\title{DECKBench: Benchmarking Multi-Agent Frameworks for\\Academic Slide Generation and Editing}

\author{Daesik Jang}
\affiliation{%
  \institution{Huawei Technologies Canada}
  \city{Burnaby}
  \state{BC}
  \country{Canada}
}
\email{daesik.jang@huawei.com}

\author{Morgan Lindsay Heisler}
\affiliation{%
  \institution{Huawei Technologies Canada}
  \city{Burnaby}
  \state{BC}
  \country{Canada}
}
\email{morgan.lindsay.heisler@huawei.com}
\orcid{0000-0003-4001-5939}

\author{Linzi Xing}
\affiliation{%
  \institution{Huawei Technologies Canada}
  \city{Burnaby}
  \state{BC}
  \country{Canada}
}
\email{linzi.xing@huawei.com}
\orcid{0009-0002-5565-3757}

\author{Yifei Li}
\affiliation{%
  \institution{University of British Columbia}
  \city{Vancouver}
  \state{BC}
  \country{Canada}
}
\email{yfli@cs.ubc.ca}

\author{Edward Wang}
\affiliation{%
  \institution{University of British Columbia}
  \city{Vancouver}
  \state{BC}
  \country{Canada}
}
\email{wanglx01@student.ubc.ca}





\author{Ying Xiong}
\affiliation{%
  \institution{Huawei Technologies Canada}
  \city{Burnaby}
  \state{BC}
  \country{Canada}
}
\email{ying.xiong2@huawei.com}

\author{Yong Zhang}
\affiliation{%
  \institution{Huawei Technologies Canada}
  \city{Burnaby}
  \state{BC}
  \country{Canada}
}
\email{yong.zhang3@huawei.com}
\orcid{0000-0002-0238-0719}

\author{Zhenan Fan}
\authornote{Corresponding Author.}
\affiliation{%
  \institution{Huawei Technologies Canada}
  \city{Burnaby}
  \state{BC}
  \country{Canada}
}
\email{zhenan.fan1@huawei.com}
\orcid{0000-0001-5116-2956}

\renewcommand{\shortauthors}{Jang et al.}

\begin{abstract}
Automatically generating and iteratively editing academic slide decks requires more than document summarization, demanding faithful content selection, coherent slide organization, layout-aware rendering, and robust multi-turn instruction following; however, existing benchmarks and evaluation protocols do not adequately measure these challenges. To address this gap, we introduce the Deck Edits \& Compliance Kit Benchmark (\texttt{DECKBench}) and evaluation framework for multi-agent slide generation and editing, built on a curated dataset of paper–slide pairs augmented with realistic, simulated editing instructions. Our evaluation protocol systematically assesses slide- and deck-level fidelity, coherence, layout quality, and multi-turn instruction-following. Furthermore, we implement a modular multi-agent baseline system that decomposes slide generation/editing task into  paper parsing and summarization, slide planning, html creation, and iterative editing. Experimental results demonstrate that our proposed benchmark can effectively highlight strengths, expose failure modes, and provide actionable insights for improving multi-agent slide generation and editing. Overall, this work establishes a standardized foundation for reproducible and comparable evaluation of academic presentation generation and editing systems. Our code and data are publicly available at \textcolor{blue}{\url{https://github.com/morgan-heisler/DeckBench}}.
\end{abstract}

\ccsdesc[500]{Computing methodologies~Multi-agent planning}
\ccsdesc[500]{Computing methodologies~Planning with abstraction and generalization}
\ccsdesc[300]{Computing methodologies~Information extraction}
\ccsdesc[300]{Computing methodologies~Natural language generation}
\ccsdesc[500]{Computing methodologies~Simulation evaluation}
\keywords{Automated presentation generation, multi-agent pipelines, evaluation benchmarks, multi-turn instruction following, structured document generation, iterative refinement, user simulation}
\begin{teaserfigure}
  \includegraphics[width=\textwidth]{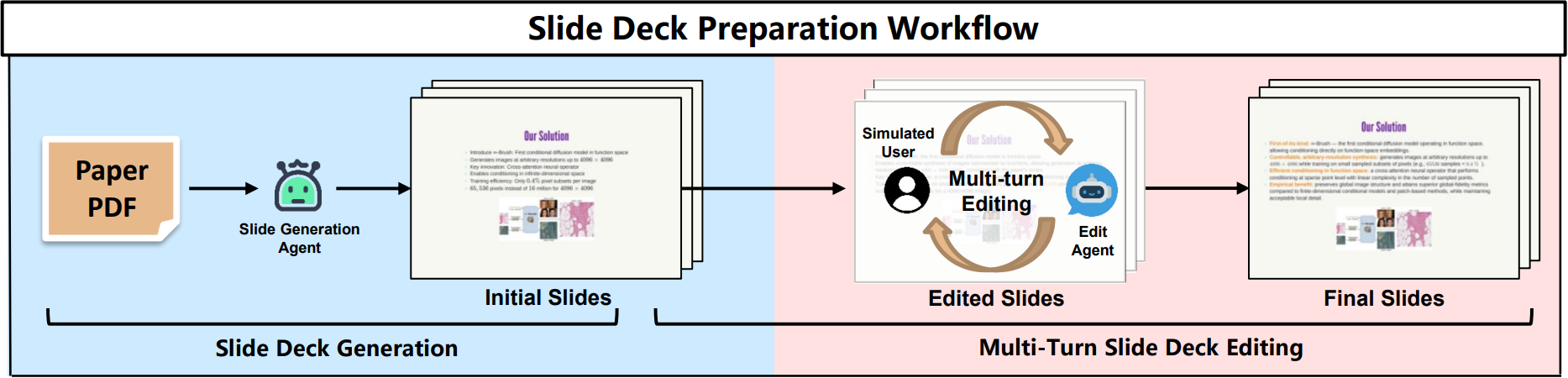}
  \caption{An end-to-end academic slide deck preparation workflow that can be thoroughly evaluated using our proposed \texttt{DECKBench}, which consists of two stages: \textcolor{blue}{[Slide Deck Generation]} — a research paper is first converted into an initial slide deck; \textcolor{pink}{[Multi-Turn Slide Deck Editing]} — a multi-turn editing loop where user instructions (from user simulated by user agent in our case) iteratively update the slides, resulting in a final presentation.}
  \Description{Enjoying the baseball game from the third-base
  seats. Ichiro Suzuki preparing to bat.}
  \label{fig:teaser}
\end{teaserfigure}

\received{8 February 2026}

\maketitle

\section{Introduction}

Preparing academic slide decks is a routine yet time-consuming part of scientific communication. Researchers regularly distill long, technical papers into concise visual presentations, revise slides in response to feedback, and adapt decks for different audiences or venues. Despite recent progress in automatic slide generation and LLM-based editing systems, existing approaches remain limited in two important ways. First, most prior work~\cite{fu2022doc2ppt,sun-etal-2021-d2s,ppsgen2013,sravanthi2009slidesgen} treats slide creation as a \textit{one-shot transformation} from text to slides, focusing primarily on summarization accuracy or static layout quality. This overlooks how presentations are produced in practice -- through \textit{iterative refinement}, where authors repeatedly edit, reorder, reformat, and augment slides over many revision cycles.


Second, while emerging multi-agent or tool-augmented LLM systems are increasingly capable of modifying existing slide decks, the field lacks a \textit{standardized benchmark} for evaluating such editing workflows. Existing datasets typically isolate either slide generation or editing, and prevailing metrics measure only static similarity to a target deck. As a result, they do not capture whether generated slides are \textit{easy to edit}, whether agents can \textit{consistently improve} a deck over multiple turns, or how well systems align with realistic user editing requests. Consequently, progress is difficult to quantify, and comparisons across generation or editing agents are often ad hoc, dataset-specific, or non-reproducible.

To address this challenge, we view \textit{user simulation} as a necessary abstraction for scalable and reproducible evaluation of multi-turn slide editing systems. Rather than relying on costly and irreproducible human-in-the-loop studies, simulated users—implemented as instruction-generating agents enable controlled evaluation of how systems respond to sequences of editing instructions. This approach allows benchmarks to focus on \textit{iterative improvement}, instruction-following behavior, and edit locality, without assuming a single “correct” final deck.

Furthermore, slides are inherently multi-modal artifacts combining text, figures, layout, transitions, and visual structure. Effective evaluation must therefore account not only for content fidelity but also for cross-slide coherence, figure selection, layout editability, and multi-turn consistency — dimensions inadequately supported by existing tools and datasets for evaluation. Meanwhile, LLM-driven slide generation systems are increasingly utilized in research labs, universities, and industry, underscoring the need for rigorous benchmarks that mirror real-world workflows.

\textbf{This gap motivates the need for a unified benchmark that evaluates both slide generation \textit{and} multi-turn editing, using simulated interaction protocols and metrics explicitly designed for iterative revision.} Such a benchmark should support multi-modal alignment, quantify iterative improvement over successive edits, and provide a reproducible framework for comparing multi-agent systems, single-step generators, and editing agents under consistent conditions. With this in mind, the contributions of this paper are as follows:


\begin{itemize}
    \item A curated dataset of academic paper-slide pairs for generation and editing.
    \item A comprehensive evaluation suite including a simulated user agent for content, coherence, and editability.
    \item Baseline results (using multi-agent systems) illuminating strengths \& limitations.
    \item Analysis of failure modes and guiding insights for future slide-generation/editing agents.
\end{itemize}

\section{Background and Related Work}

Automating the transformation of scientific documents into presentation formats has become an increasingly active research area, driven by advances in large language models and multi-modal agent systems. Early research framed slide generation primarily as a summarization task. Systems such as SlidesGen \cite{sravanthi2009slidesgen} and its section-based extension SlideGen \cite{sefid2021slidegen} demonstrated that mapping paper structure to slide structure is feasible using abstractive techniques, while works like PPSGen \cite{ppsgen2013} and phrase-based methods for slide generation \cite{wang2017phrase} leveraged supervised mappings from textual fragments to slide content. DOC2PPT \cite{fu2022doc2ppt} extended this paradigm by performing end-to-end scientific-document-to-presentation generation, illustrating the potential for multi-modal template filling beyond pure summarization.

More recent efforts have expanded the scope of scientific communication beyond slide decks. Paper2Poster \cite{pang2025paper2poster} targets structured poster layouts, Paper2Video \cite{zhu2025paper2videoautomaticvideogeneration} and PresentAgent \cite{shi2025presentagent} generate video presentations, and Paper2SysArch \cite{guo2025paper2sysarch} produces constrained system-architecture diagrams. These works collectively show that scientific papers can serve as rich input for multi-modal generation pipelines. However, evaluation practices remain fragmented, with each system introducing its own metrics and only narrow modality-specific benchmarks.

A parallel research thread focuses on interactive or agentic systems for presentation creation. Multi-agent frameworks such as AutoPresent \cite{autopresent2025}, Auto-Slides \cite{yang2025auto}, PreGenie \cite{xu2025pregenie}, PPTAgent \cite{zheng2025pptagent}, and PresentCoach \cite{chen2025presentcoach} decompose slide creation into coordinated retrieval, summarization, layout, and revision modules. The newly introduced SlideGen \cite{liang2025slidegencollaborativemultimodalagents} advances this direction significantly by proposing a design-aware, multi-modal, vision-in-the-loop multi-agent system that jointly performs outlining, semantic mapping, visual arrangement, and iterative refinement. Unlike earlier summarization-heavy approaches, SlideGen emphasizes deliberate visual planning and coordinated cross-agent reasoning, achieving state-of-the-art results in visual quality, layout organization, and content faithfulness. Several works incorporate human feedback loops, including Slide4N \cite{wang2023slide4n}, which supports human-AI collaboration in notebook-derived slides, and Human-Agent Collaborative Paper-to-Page Crafting \cite{ma2025human}, which emphasizes ultra-low-cost iterative refinement. Additional systems address personalization and pedagogical adaptation, as seen in persona-aware presentation generation \cite{mondal2024presentations} and situated-teaching-oriented slide adaptation \cite{liu2025addressing}. These works collectively highlight a shift from static generation to interactive workflows in which editing capability and iterative improvement are central. Despite this shift, there is still no standardized evaluation protocol for multi-turn slide editing, nor any benchmark that systematically assesses editing responsiveness and revision quality at scale.

Complementing these generation systems, a growing body of work seeks to evaluate LLM or VLM capabilities within presentation environments. The PPTC \cite{guo2024pptc} and PPTC-R \cite{zhang2024pptc} benchmarks introduced structured task suites covering task completion, robustness, and layout manipulation in PowerPoint interfaces. VLM-SlideEval \cite{kang2025vlm} extended this line by testing structured comprehension and perturbation sensitivity for visual-linguistic slide understanding. While these benchmarks represent important progress, they are limited to atomic tasks or diagnostic robustness checks. They do not measure holistic paper-to-slide generation quality, nor do they evaluate multi-turn editing competence, leaving a gap in unified benchmarks for end-to-end presentation workflows.

More recently, systems such as Talk to Your Slides \cite{jung2025talkslideslanguagedrivenagents} and PPTAgent \cite{zheng2025pptagent} demonstrate the feasibility of language-driven slide editing with fine-grained multi-modal control. These efforts underscore the importance of treating editing as a first-class reasoning task. Nevertheless, the field lacks a comprehensive benchmark that evaluates (1) fidelity of paper-to-slide generation, (2) editability and multi-turn responsiveness, and (3) overall end-to-end agentic performance under realistic usage patterns. Without such a unified framework, it remains difficult to compare models or measure how summarization, multi-modal reasoning, layout organization, and editing capabilities interact in practical slide-creation settings.

To address these gaps, we introduce a benchmark for academic-paper-to-slide generation paired with a simulated user-editing framework that reflects real, multi-turn revision cycles. Our benchmark unifies evaluation across content fidelity, structural alignment, visual organization, multi-modal reasoning, and editability. By situating slide editing within the broader scientific communication workflow, from raw paper to refined presentation, our work provides the first comprehensive testbed for evaluating end-to-end performance of LLM-powered presentation agents.

\section{Design of the Benchmark Dataset}
\label{sec:dataset}

We introduce the \texttt{DECKBench} (Deck Edits \& Compliance Kit) Benchmark.  

\subsection{Scope and Use Cases}

The benchmark targets two complementary tasks that reflect the end-to-end workflow of preparing academic research presentations. The first task evaluates slide generation, where an agentic system receives a full academic paper and produces a slide deck that communicates the paper's core ideas, methodology, and empirical results. This setting captures the challenges of long-context comprehension, multi-modal summarization, and presentation structuring.

The second task evaluates slide editing. Here, the agentic system is given an existing slide deck along with a natural-language editing request, and must output a revised deck that incorporates the requested changes while preserving structural coherence. To accurately reflect real usage patterns, we frame slide editing as an iterative process rather than a single transformation step. This corresponds to a realistic scenario in which a user continually refines their slides, issuing edits as they compare the current deck with the intended final version.

\subsection{Data Collection and Curation}
\label{sec:datacuration}

We construct an evaluation benchmark of paired academic papers and presentation slides drawn from four major machine learning and computer vision venues: CVPR, ECCV, ICLR, and ICML. The corpus spans multiple conference years and captures substantial variation in topic, presentation style, document length, figure density, and mathematical content. For each paper, we retrieve the corresponding official presentation slides and retain only paper--slide pairs for which both artifacts are available and parsable, yielding \textbf{294} matched pairs.

\subsection{Simulated User Editing }

\begin{figure}[htbp!]
  \vskip 0.2in
  \begin{center}
    \centerline{\includegraphics[width=\columnwidth]{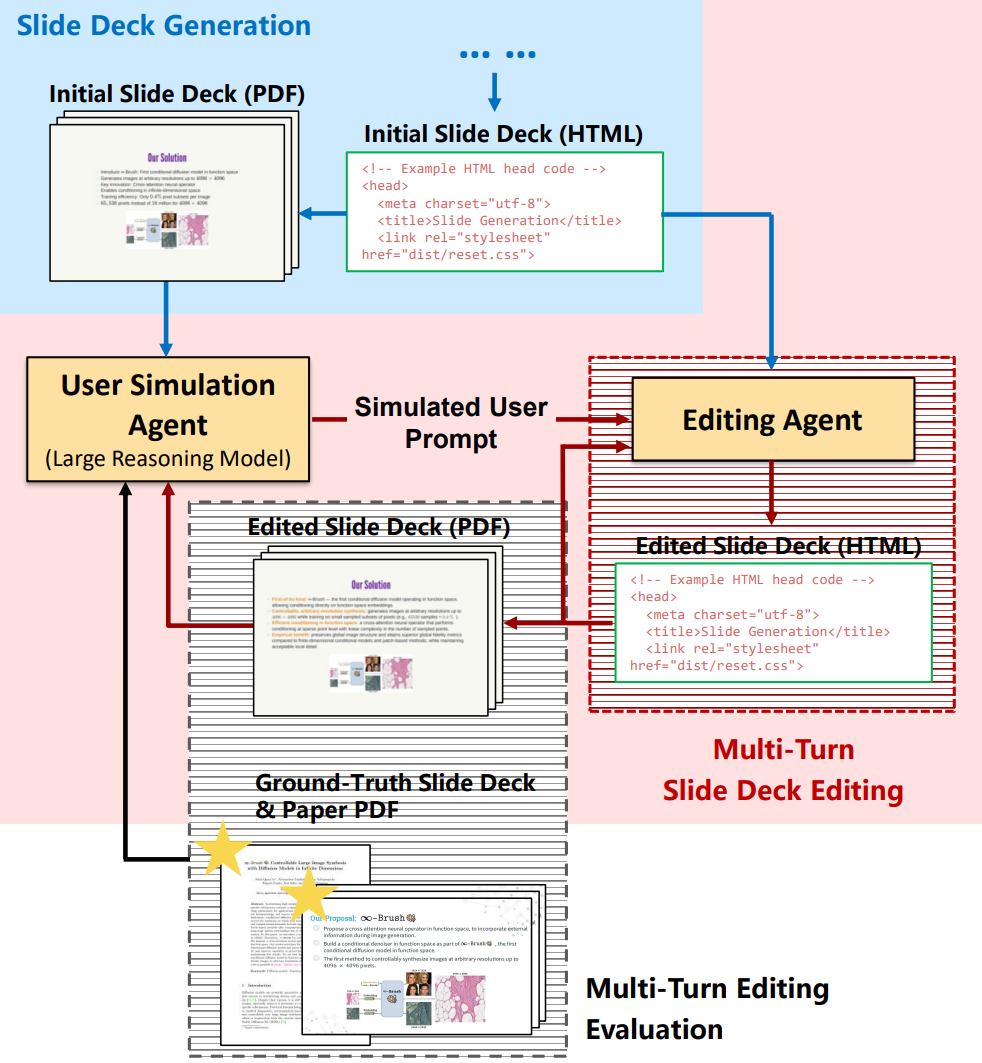}}
    \caption{
      Overview of the multi-turn slide editing evaluation pipeline. A user simulation agent interacts with the editing agent to iteratively refine the HTML slide deck, with each iteration evaluated against ground-truth decks using metrics from Section~\ref{sec:multi-turn-eval}. The complete slide generation pipeline (Fig.~\ref{fig:slide-overview}) is omitted for brevity.
    }
    \Description{Overview of the User Simulation and Editing Pipeline. The User Simulation agent emulates a real human by creating a simulated editing prompt which is then fed to the editing agent, along with the initial slide deck. Then the editing agent edits the initial HTML slide deck and generates a new HTML slide deck. This can then be used as an input the user simulation agent to create more iterations of editing.}
    \label{fig:editing-overview}
  \end{center}
\end{figure}

We adopt a \textit{simulated user} paradigm inspired by recent interactive evaluation frameworks (e.g., $\tau$$^2$-Bench~\cite{barres2025tau2benchevaluatingconversationalagents}). In contrast to traditional editing datasets that rely on hand-crafted or crowdsourced gold-standard edit requests, our approach uses an automated \textit{user simulator} to generate editing instructions. For each paper–slide pair, the simulator compares the current intermediate slide deck with the corresponding ground-truth final deck and derives an editing request that approximates how a real user might describe the changes needed to move the slides toward the desired end state. This process is shown in Figure~\ref{fig:editing-overview}.

To better reflect the diversity of real-world user behavior, the simulator supports three distinct user personas that differ in communication style and specificity. These personas span highly verbose users who provide justification and contextual detail, as well as concise users who issue short, directive instructions focused on surface-level edits. Persona-specific prompting controls linguistic granularity, edit framing, and verbosity, allowing us to evaluate model robustness under varied simulated interaction styles rather than optimizing for a single canonical instruction format.

Using this simulator, we generate 5 multi-turn editing interactions across the dataset. Each turn consists of an intermediate slide deck state paired with an automatically generated editing instruction. This setup enables fine-grained analysis of stepwise editing behavior, instruction-following fidelity, and multi-turn consistency, while explicitly modeling user interaction through simulation.


\section{Evaluation Protocol and Metrics}
\label{sec:metrics}
We evaluate systems at three complementary levels: slide-level, deck-level, and multi-turn interaction-level. Each level, as illustrated in Table~\ref{tab:metrics-overview}, includes both reference-free and reference-based metrics, capturing different aspects of linguistic quality, content fidelity, structural coherence, and interactive performance. Further technical details for each metric are expanded upon in the remainder of this section.


\begin{table*}[t]
\centering
\caption{Hierarchical overview of evaluation metrics designed in DeckBench, covering slide generation quality and multi-turn editing effectiveness. Reference-free and reference-based metrics are colorcoded in \textcolor{blue}{blue} and \textcolor{red}{red} respectively.}
\label{tab:metrics-overview}
\begin{tabular}{lll}
\toprule
\textbf{Metric Category} & \textbf{Subcategory} & \textbf{Metrics} \\
\midrule

\multirow{4}{*}{Slide Generation Metrics}
& \multirow{2}{*}{Slide Level}
& \textcolor{blue}{Perplexity}, \textcolor{blue}{Faithfulness}, \textcolor{red}{Text Similarity},\\
& & \textcolor{red}{Figure Similarity, Layout Quality} \\

\cmidrule(lr){2-3}

& \multirow{2}{*}{Deck Level}
& \textcolor{blue}{Perplexity}, \textcolor{blue}{Faithfulness}, \textcolor{blue}{Fidelity}\\
& & \textcolor{red}{Transition Similarity}, \textcolor{red}{Dynamic Time Warping (DTW)} \\

\midrule

Multi-Turn Editing Metrics
&
& \textcolor{red}{$\Delta$DTW}, \textcolor{red}{$\Delta$Transition Similarity} \\

\bottomrule
\end{tabular}
\end{table*}


\subsection{Slide-Level Metrics}

Because generative systems may produce a different number of slides and reorder content, we first align generated slides with their ground-truth counterparts before computing any slide-level metrics. We perform alignment using a Hungarian matching procedure. For each pair of generated and reference slides, we compute a similarity score based on combined textual and visual features, and then solve for the optimal one-to-one assignment that maximizes total similarity. This alignment ensures fair evaluation even when slide counts or ordering differ.

\paragraph{Slide-Level Evaluation Overview}
We group our slide-level evaluation metrics into two broad classes: \textit{reference-based} metrics, which measure similarity to the ground-truth slide deck after alignment, and \textit{reference-free} metrics, which assess intrinsic quality and grounding without relying on human-authored slides.

\paragraph{Reference-Based Metrics}
Reference-based evaluation assesses alignment with the human-authored slide deck after Hungarian matching. Textual similarity is computed as the cosine similarity between embedding representations of each generated slide and its matched reference slide. For visual elements such as figures, we use a vision–language model (CLIP-ViT-base-patch32~\cite{radford2021learningtransferablevisualmodels}) to compute similarity between generated and reference figures, capturing whether the generated slide appropriately reflects the intended visual content.

\paragraph{Reference-Free Metrics}
Reference-free evaluation focuses on the linguistic quality and semantic grounding of individual slides. We first measure fluency and coherence using perplexity (PPL) computed over the slide text. Let $x_1, x_2, \dots, x_n$ denote the sequence of tokens in a slide. Perplexity is defined as:
\[
\mathrm{PPL} = \exp\Biggl(- \frac{1}{n} \sum_{i=1}^{n} \log P(x_i \mid x_1, \dots, x_{i-1}) \Biggr),
\]
where $P(x_i \mid x_1, \dots, x_{i-1})$ is the token probability predicted by a language model.

In addition to simple language quality, we evaluate semantic grounding with respect to the source paper. Although this metric does not reference the human-produced slides, it leverages the original paper as an external source of truth. For each generated slide $s$, we identify the most semantically similar paper chunk $p^{*}$ using embedding similarity, and compute the cosine similarity between $s$ and $p^{*}$ as a faithfulness score. This captures whether slide content is meaningfully grounded in the paper rather than hallucinated.

\subsection{Deck-Level Metrics}

At the deck level, the evaluation emphasizes global coherence, coverage, and narrative structure of generated slide decks. \textit{Reference-based} deck metrics assess structural alignment with human slides. Dynamic Time Warping (DTW)~\cite{1163055} computes a soft alignment of slide sequences, allowing decks of different lengths or non-monotonic orderings to be compared. Transition similarity further evaluates whether the progression of slide embeddings in the generated deck mirrors the reference sequence. Finally, an LLM-as-a-judge can be applied in a reference-aware setting, scoring content completeness, faithfulness, and alignment with canonical presentation structure.

\textit{Reference-free} metrics include deck-level perplexity, which measures fluency across the concatenated text of all slides. Deck fidelity quantifies coverage of the paper’s content: each paper chunk is aligned with its most similar slide, and the fidelity score is the average similarity over all paper chunks. Deck faithfulness, in contrast, evaluates the grounding of each slide in the source paper by averaging similarity scores across all slide-paper pairs. To capture higher-level narrative quality, we also employ an LLM-as-a-judge approach. A large language model evaluates the coherence of slide transitions, logical flow, ordering, and summarization quality, producing a holistic score for the deck. 

\subsection{Multi-Turn Evaluation (Slide Editing / Interactive Agents)}
\label{sec:multi-turn-eval}
For interactive or multi-turn evaluation, the benchmark measures a system’s ability to iteratively revise a slide deck in response to simulated user instructions. 

Reference-based multi-turn metrics measure progress toward the ground-truth deck at each editing step. The per-turn improvement in DTW distance, $\Delta \mathrm{DTW}_t = \mathrm{DTW}(s_{t-1}, s^*) - \mathrm{DTW}(s_t, s^*)$, indicates how much closer the generated deck $s_{t-1}$ moves toward the human-edited reference $s^*$ after applying the user instruction to become the edited deck $s_{t}$. Similarly, per-turn transition similarity improvement measures the alignment of slide-to-slide embeddings relative to the reference deck, quantifying how effectively the system converges toward the intended narrative structure over multiple editing iterations.

Taken together, these metrics provide a comprehensive framework for evaluating both static generation and interactive slide-editing capabilities, combining low-level linguistic and visual fidelity with high-level structural and iterative performance.

\section{Multi-Agent Baseline System Architecture}
\label{sec:baseline_}
\begin{figure*}[ht]
  \vskip 0.2in
  \begin{center}
    \centerline{\includegraphics[width=\textwidth]{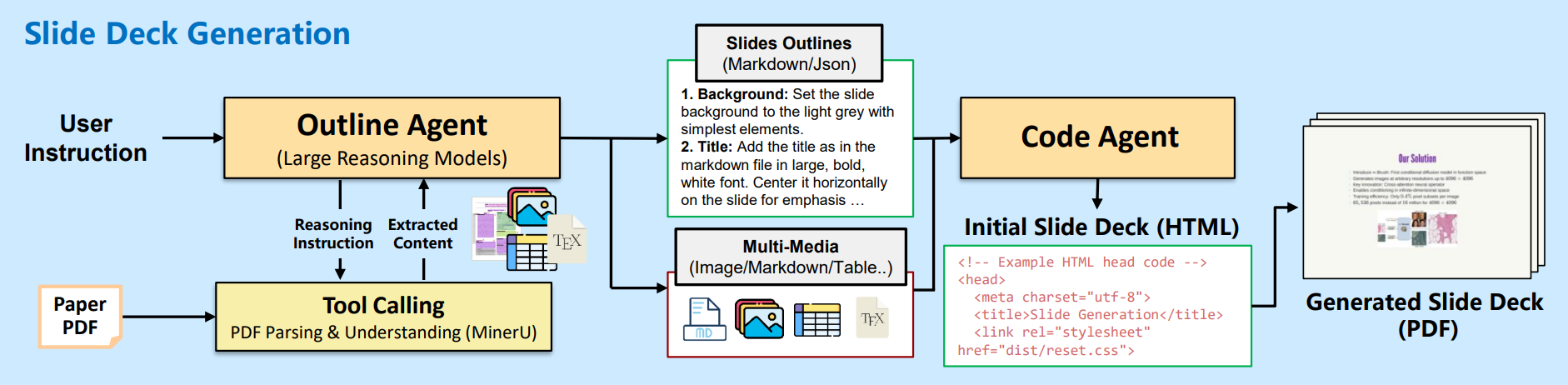}}
    \caption{
      Overview of the slide generation pipeline (in Section~\ref{sec:baseline_}). User instructions are processed by an outline agent that extracts key paper content and generates a slide outline, which is then converted by a code agent into compilable HTML slides.}
    \Description{This figure shows a diagram of the system architecture. On a high level, the user instruction goes into the outline agent which creates an outline of the slides by reasoning on the user preference and call information extraction tool to extract necessary information from the original paper, which is then fed into the code agent which create the compilable HTML code for the generated slide deck.}
    \label{fig:slide-overview}
  \end{center}
\end{figure*}

Coupled with DeckBench, we also propose a reference multi-agent system designed to transform scientific papers into editable presentation decks and to support subsequent multi-turn editing. The architecture, as illustrated in Figure~\ref{fig:slide-overview}, follows a modular pipeline in which each agent specializes in a distinct stage of the workflow. Although simple by design, this baseline establishes a consistent evaluation target and exposes the practical challenges faced by end-to-end slide generation systems, including document parsing, content selection, layout generation, and iterative refinement.

The pipeline begins with the \textbf{Outline Agent}, whose role is to convert a PDF source document into a structured representation of the paper. The agent delegates low-level parsing to the MinerU toolkit \cite{he2024opendatalab, wang2024mineruopensourcesolutionprecise, niu2025mineru25decoupledvisionlanguagemodel}, which reliably extracts the document’s hierarchy, including sections, headings, figures, tables, equations, and salient sentences. Based on these raw elements, the agent produces an intermediate outline that contains summaries of each section along with candidate bullet points intended for slides. This stage performs the highest-level content planning in the system: it determines what information is preserved, what is condensed, and how content should be distributed across the eventual deck.

The system then forwards the extracted outline to the \textbf{Code Agent} whose objective is to construct an initial slide deck in an editable HTML-based format. The agent maps outline elements to slide templates, assigns titles and bullet lists, and inserts figures or tables when available. This component also handles the conversion of the slide representation into a format suitable for downstream editing and rendering. Because layout generation is not learned in this baseline, the Code Agent relies on a pre-defined CSS template for style-related items such as font choice and colour. This design choice isolates the effects of content selection from the complexity of layout modeling, providing a clean foundation for benchmarking methods that incorporate richer visual reasoning.

Once a slide deck is produced by the Code Agent, a single \textbf{Editor Agent} supports iterative refinement through natural-language instructions. Given the current slide deck and a user instruction, the Editor Agent performs a unified editing action that jointly accounts for the user’s intent, the relevant portion(s) of the slide deck, and the concrete modifications required to the slide HTML. These aspects are handled within one end-to-end editing step rather than as separate tasks or modules.

The Editor Agent supports a broad range of edits, including rewriting bullet points, adding or removing figures, reordering slides, and improving visual clarity. This design reflects realistic presentation-authoring workflows, in which users iteratively refine an existing deck through many small, semantically grounded adjustments instead of regenerating the entire presentation from scratch.

The entire architecture design of our deck generation and editing system incorporates several implementation decisions motivated by robustness and scalability. The Outline and Editor Agents rely on \textit{thinking} Large Reasoning Models (LRMs) for high-level planning, summarization, content selection, and instruction interpretation, whereas the Code Agent relies on \textit{coding} LLMs to ensure compilable HTML outputs. Extracting figures and tables introduces practical tradeoffs: while the system attempts to place all relevant visual elements referenced in the outline, missing metadata or complex figure arrangements can cause placement errors, which the Editor Agent must later correct. Long papers pose additional challenges because structured extraction produces large outlines that must be compressed without losing key contributions. To handle such cases, the Outline Agent uses hierarchical summarization, and the system prompt can be modified to include the intended number of slides per deck.

Finally, the baseline includes fallbacks for parsing failures, layout inconsistencies, or conflicting edit operations. When parsing yields incomplete document structure, the Outline Agent defaults to section-level text extraction. The pipeline also supports human-in-the-loop oversight for real-world applications: a user may inspect automatically generated slides, approve or revise them, and continue iterative editing. These mechanisms ensure that the multi-agent baseline remains functional across the wide variety of document formats and editing styles encountered in practice.

\section{Experimental Setup}

This section presents the experimental setup used to evaluate our proposed benchmark (Sec.~\ref{sec:dataset}) and baseline system (Sec.~\ref{sec:baseline_}).

\subsection{Tasks}
We evaluate systems on two complementary tasks:\par\noindent
\textbf{[T1] Paper-to-Slide Generation:}
Given a full academic paper $P$, the system generates an initial slide deck $S_0$. This task measures long-context understanding, summarization ability, visual organization, and structural coherence.\par\noindent
\textbf{[T2] Multi-Turn Iterative Editing:}
Given an initial deck $S_0$ and an editing instruction $\mathcal{I}_1$ sampled from one of our simulated user personas, systems must output an edited deck $S_1$. This continues for a sequence of $T$ user instructions ${\mathcal{I}_1, \dots, \mathcal{I}_T}$, so the systems revise the deck iteratively to obtain $S_T$. This setting reflects realistic revision cycles for preparing academic presentations.


Across all tasks, systems are evaluated with both reference-free and reference-based metrics proposed in Sec.~\ref{sec:metrics}, capturing alignment to the user instruction stream and convergence toward the human-produced final slide deck.

\subsection{Dataset Statistics}

We evaluate models on the \texttt{DECKBench} dataset introduced in Section~\ref{sec:dataset} and Table~\ref{tab:dataset-stats} reports the full statistics of this dataset. Note that math density is estimated using a heuristic that counts math-heavy text lines containing symbolic operators or Greek letters. This metric reflects relative mathematical content rather than exact equation counts. Additionally, slide refers to a single presentation slide, slide deck refers to the whole deck of presentation slides, and paper refers to the academic paper.

\begin{table}[ht]
\centering
\caption{Statistics of the \texttt{DECKBench} dataset.}
\label{tab:dataset-stats}
\begin{tabular}{lc}
\toprule
\# Paper-Deck Pairs & 294 \\
Avg. Paper Length (pages) & 19.5 ± 9.4 \\
Avg. Slides / Deck & 15.4 ± 8.2 \\
Avg. Figures / Paper & 8.3 ± 19.5 \\
Avg. Figures / Slide Deck & 47.4 ± 46.5 \\
Avg. Math Density / Paper & 62.8 ± 74.2 \\
Avg. Math Density / Slide Deck & 5.7 ± 14.6 \\
\# Persona Types & 3 \\
\bottomrule
\end{tabular}
\end{table}

\subsection{Systems Compared}

We evaluate the proposed pipeline (Sec.~\ref{sec:baseline_}) and compare it against other previously proposed multi‑agent baselines. All of which operationalize paper‑to‑slide generation through coordinated reasoning across specialized modules, but they differ in the specific language and vision models that underpin agent capabilities. \textbf{Auto‑Slides}~\cite{yang2025auto} leverages large‑scale language models accessed via API (by default OpenAI’s GPT‑4o family) for document comprehension, content summarization, and slide text generation, with additional auxiliary models such as marker‑pdf for PDF parsing and structure extraction. \textbf{EvoPresent}~\cite{liu2025presentingpaperartselfimprovement} integrates a self‑improvement aesthetic model, PresAesth, trained via multi‑task reinforcement learning on aesthetic preferences (built on a Qwen2.5‑VL‑7B backbone for visual evaluation), while text and visual content extraction and generation agents may employ a mix of pre-trained LLMs depending on the pipeline stage. As the trained model was not open-source at the time of writing, we used GPT-4o for all models in this pipeline. \textbf{SlideGen}~\cite{liang2025slidegencollaborativemultimodalagents} organizes a cohort of vision-language agents that reason collaboratively over document structure and semantics; these agents are implemented atop multi-modal large language models capable of jointly processing text and visual inputs to coordinate outlining, mapping, layout design, note synthesis, and refinement. 
These differences in underlying model families and multi-modal capabilities directly affect each baseline’s capacity for reasoning over long contexts, aligning visual assets, and managing layout generation, motivating their inclusion as comparative systems for complex scientific slide generation.




\subsection{Implementation}


\textbf{LLMs.}
We experimented with multiple model families. For the Qwen results, the Outline Agent used Qwen3-30B-A3B-Thinking-2507 and for the Code Agent, Qwen2.5-Coder-14B-Instruct-AWQ was used. Long-context inputs are chunked via MinerU PDF parsing. We tested two combinations of user simulation model and editor model: GPT-5.1 and GPT-5-mini, as well as Kimi-K2~\cite{kimiteam2026kimik2openagentic} and DeepSeek-V3.1~\cite{deepseekai2024deepseekv3technicalreport}.

\noindent\textbf{HTML Slide Renderer.}
We use a lightweight HTML-based slide format compatible with Reveal.js. All decks are rendered at 1080p for evaluation.


\begin{table*}[htbp]
\centering
\caption{Task 1 (Paper-to-Slide Generation) results for Slide Metrics on DECKBench.}
\label{tab:generation-main}
\begin{tabular}{lcccccc}
\toprule
 &  & \multicolumn{2}{c}{\textcolor{blue}{Reference Based}}  & \multicolumn{3}{c}{\textcolor{red}{Reference Free}}  \\
\cmidrule{3-7}
System & Model & \textcolor{blue}{PPL}$\downarrow$ & \textcolor{blue}{Faith.}$\uparrow$ & \textcolor{red}{TextSim}$\uparrow$ & \textcolor{red}{FigSim}$\uparrow$ & \textcolor{red}{Layout}$\uparrow$ \\
\midrule
Evo-Present & GPT-4o & 184.183 & 0.408 & 0.567 & 0.535 & 0.696\\
SlideGen & GPT-4o & 208.330 & 0.535 & 0.673 & \textbf{0.681} & 0.960\\
Auto-Slides & GPT-4o & \textbf{90.528} & 0.534 & \textbf{0.729} & 0.465 & 0.998 \\
\midrule

Ours & GPT-4o & 191.731 & \textbf{0.538} & 0.698 & 0.590 & \textbf{1.000}\\
Ours & Qwen & 134.411 & 0.531 & 0.700 & 0.591 & 0.999\\
\bottomrule
\end{tabular}
\end{table*}

\begin{table*}[htbp]
\centering
\caption{Task 1 (Paper-to-Slide Generation) results for Deck Metrics on DECKBench.}
\label{tab:generation-main-deck}
\begin{tabular}{lccccccc}
\toprule
 &  & \multicolumn{3}{c}{\textcolor{blue}{Reference Based}}  & \multicolumn{2}{c}{\textcolor{red}{Reference Free}} &  \\
\cmidrule{3-7}
System & Model & \textcolor{blue}{PPL}$\downarrow$ & \textcolor{blue}{Faith.}$\uparrow$ & \textcolor{blue}{Fidelity}$\uparrow$ & \textcolor{red}{DTW}$\uparrow$ & \textcolor{red}{TransSim}$\uparrow$& Fail Rate$\downarrow$\\
\midrule
Evo-Present & GPT-4o & 29.769 & 0.546 & 0.528 & 0.759 & 0.819&0.000 \\
SlideGen & GPT-4o &  30.742 & 0.659 & 0.566 & 0.772 & 0.856 &0.014\\
Auto-Slides & GPT-4o &  \textbf{14.982} & 0.691 & \textbf{0.591} & \textbf{0.783} & 0.858 &0.078\\
\midrule

Ours & GPT-4o & 45.646 & 0.691 & 0.584 & 0.773 & \textbf{0.879} & 0.000 \\
Ours & Qwen & 28.220 & \textbf{0.693} & 0.588 & 0.772 & 0.875 & 0.000 \\
\bottomrule
\end{tabular}
\end{table*}

\section{Results}

We now present results for the two tasks: paper-to-slide generation and multi-turn iterative editing. 

\subsection{Paper-to-Slide Generation}

Tables~\ref{tab:generation-main} and \ref{tab:generation-main-deck} compare our method against existing paper-to-slide generation systems on \texttt{DECKBench}. Auto-Slides achieves the highest faithfulness and slide fidelity scores, indicating strong content preservation, but underperforms on figure alignment and exhibits a substantially higher failure rate (7.8\%). In contrast, our approach maintains competitive faithfulness while improving figure alignment and layout consistency, achieving perfect or near-perfect reliability (0\% failure). These results suggest that, while no system dominates all metrics, our method provides a more balanced and robust solution for paper-to-slide generation. 

\subsection{Multi-Turn Iterative Editing}

\begin{figure}[ht]
\centering

\begin{subfigure}[b]{\columnwidth}
  \includegraphics[width=1\linewidth]{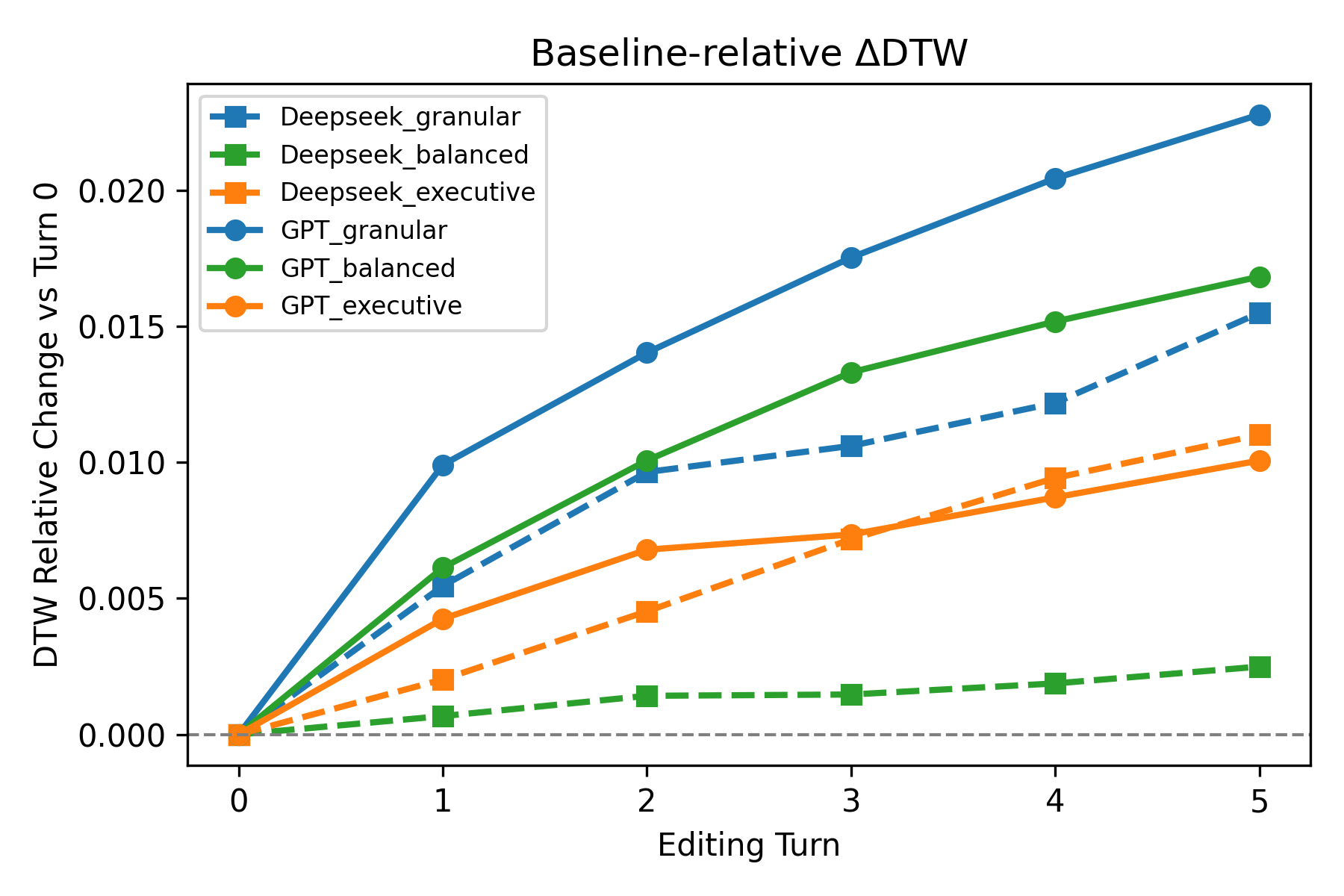}
  \caption{}
  \label{fig:persona_effect1} 
\end{subfigure}

\medskip 
\begin{subfigure}[b]{\columnwidth}
  \includegraphics[width=1\linewidth]{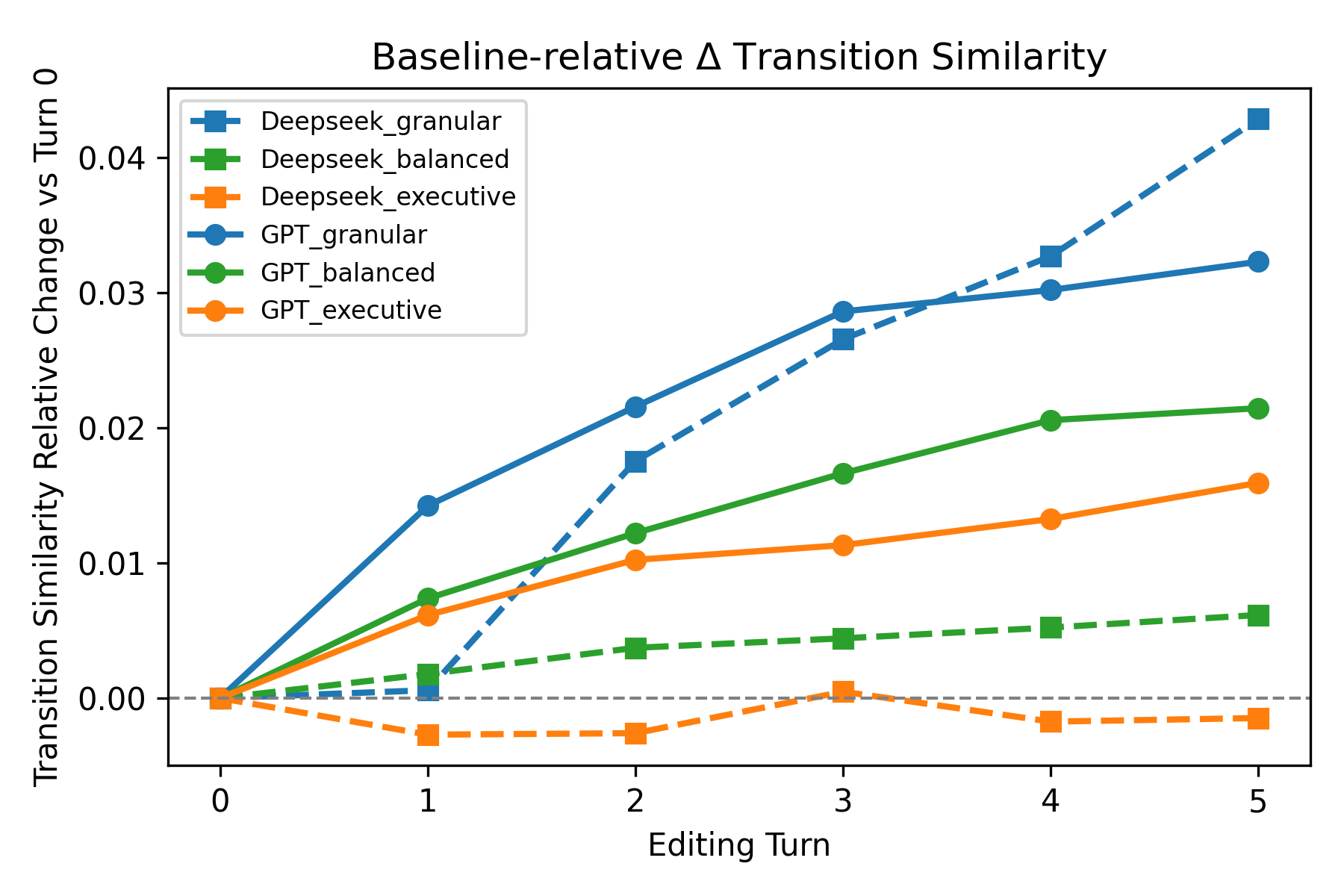}
  \caption{}
  \label{fig:persona_effect2}
\end{subfigure}

\caption[Two numerical solutions]{%
(a) Baseline-relative $\Delta$DTW for different models and personas. 
(b) Baseline-relative $\Delta$ Transition Similarity for different models and personas.}
\label{fig:persona_effect}
\Description{This is a line plot that shows the DTW and transition similarity trend per editing turn. From the plot, we can see that both metrics increase with each editing turn. }
\end{figure}

We analyze multi-turn iterative editing by measuring cumulative relative changes in similarity metrics with respect to the initial response (Turn~0), summarized after five editing turns in Table~\ref{tab:multiturn}. We analyze across the three personas (granular, balanced, executive) which correspond to high, medium and low granularity respectively. Examples of prompts from these personas are provided in Appendix~\ref{appendix:prompts}. Across all evaluated settings, DTW distance increases monotonically with turn index, indicating steady structural drift as iterative edits accumulate. In contrast, transition similarity exhibits more heterogeneous behavior across models and interaction granularities. As seen in Figure~\ref{fig:persona_effect}, GPT-based settings show substantial and consistent increases in transition similarity over turns—particularly under granular prompting—while Kimi~K2-DeepSeek-based configurations exhibit markedly smaller changes, and in the executive setting, near-zero or slightly negative transition shifts during early turns. For example, under Kimi~K2-DeepSeek–balanced, $\Delta$DTW reaches 0.00249 after five turns while $\Delta$TransSim remains limited at 0.00615, whereas GPT–granular attains substantially larger changes in both metrics ($\Delta$DTW = 0.02278, $\Delta$TransSim = 0.03230). Across conditions, DTW evolves more smoothly and with lower variance than transition similarity, which displays higher sensitivity to editing strategy and prompt granularity. These results suggest that structural divergence is a reliable consequence of multi-turn editing, while semantic transition changes are more contingent on interaction design, and may stabilize or fluctuate rather than uniformly increase.



\begin{table}[htbp!]
\centering
\caption{Task 2 (Multi-turn iterative editing) performance averaged after $5$ turns.}
\label{tab:multiturn}
\begin{tabular}{lcccc}
\toprule
UserSim & Editor & Granularity & $\Delta$DTW$\uparrow$ & $\Delta$TransSim$\uparrow$ \\
\midrule
GPT-5.1 & GPT-5-mini & Low &0.01006 & 0.01594 \\
GPT-5.1 & GPT-5-mini & Medium &0.01682 & 0.02145 \\
GPT-5.1 & GPT-5-mini & High & 0.02278 & 0.03230 \\
\hline
Kimi~K2&DeepSeek  & Low & 0.01101 & -0.00147 \\
Kimi K2&DeepSeek  & Medium & 0.00249 & 0.00615 \\
Kimi K2&DeepSeek  & High & 0.01549 & 0.04287 \\
\bottomrule
\end{tabular}
\end{table}

\subsection{Qualitative Results}
We qualitatively examine generated slide decks to characterize common strengths, failure modes, and the effectiveness of targeted interventions beyond aggregate metrics.

\subsubsection{Deck Generation Examples.}
High-quality generations across systems demonstrate faithful content selection, concise bulletization, and reasonable alignment between figures and surrounding text. In contrast, low-quality examples frequently suffer from two types of degradation: layout and content.

All evaluated systems exhibit elements extending beyond slide boundaries, and most struggle to select appropriate figure scales, resulting in either illegible or visually dominant images. Additionally, EvoPresent is particularly prone to overlapping elements.

Distinct content-level failures are observed across systems. AutoPresent frequently produces incorrect figure captions, with a substantial fraction rendered in Chinese regardless of source language, indicating systematic caption localization errors. EvoPresent often substitutes placeholders or generic equations for paper-specific content, reducing faithfulness. Our system avoids hallucinated equations but remains sensitive to upstream errors, such as incorrect translation from markdown to HTML or incorrect figure paths, that can propagate into presentation artifacts.

\subsubsection{Multi-Turn Revisions.}
Multi-turn revision results indicate that iterative editing substantially improves deck quality. Across systems, successive revisions reduce layout overflow and correct many figure placement errors, with convergence typically occurring within a small number of edit steps. Notably, our system demonstrates faster convergence to visually coherent decks, as later revisions focus primarily on stylistic refinement rather than structural correction, suggesting that strong first-pass layouts reduce downstream editing burden.

\section{Future Work \& Directions}
Several avenues remain to further advance the capabilities and usability of automated slide generation systems. First, improving multi-agent pipelines presents a promising direction, particularly in enhancing figure and table understanding, optimizing slide layout, handling complex mathematical expressions and equations, and achieving robust cross-domain generalization. Second, there is a need for more rigorous, user-centered evaluation, including systematic usability studies, assessment of audience comprehension, and evaluation of accessibility for diverse user populations. Third, extending the current benchmark to encompass full video generation, including synchronized speaker audio, video, and subtitles, or interactive slide decks, would more comprehensively evaluate system performance in realistic presentation scenarios. Fourth, supporting edit history, collaborative editing, and version control would facilitate iterative refinement and team workflows. Finally, incorporating multilingual support, accessibility features, and the ability to customize outputs according to presentation style, such as lectures, posters, or conference talks, would broaden the applicability of these systems and increase their real-world impact. Collectively, these directions aim to enhance both the technical robustness and practical utility of automated presentation tools.

\section{Conclusion}
We presented \texttt{DECKBench}, a benchmark for academic slide generation and multi-turn editing, featuring diverse paper-slide pairs and a comprehensive suite of metrics for slide-, deck-, and interaction-level evaluation. We introduced a multi-agent baseline that demonstrates the challenges of parsing, content selection, layout, and iterative refinement. Our results highlight the need for standardized, reproducible evaluation in this emerging area. By releasing the dataset, metrics, and baseline, we aim to enable fair comparisons, accelerate research, and inspire future advances in interactive, multi-modal scientific presentation generation.

\section*{Ethical Statement}

Potential societal impacts are largely positive, as it can reduce the time and effort required for researchers to prepare high-quality presentations, democratize access to presentation tools, and support education and knowledge dissemination. However, automated slide generation and editing could also be misused to create superficially polished presentations that misrepresent research findings if not used responsibly. We emphasize that our benchmark encourages alignment with source content and multi-turn fidelity, mitigating risks of miscommunication. 



\bibliographystyle{ACM-Reference-Format}
\balance
\bibliography{software}

\appendix
\onecolumn

\section{Personas Used in Our Experiments}
\label{appendix:prompts}

For transparency and reproducibility, we include the exact personas used in our experiments. The personas appear below verbatim.

\subsection{Persona A: Granular Analyst}
{\small
\begin{verbatim}
        "prompt_verbosity": "detailed",
        "prompt_detail_level": "generate prompt with explicit instructions including the slide titles, bullet points, 
        table rows, captions, and example numbers necessary for replicating the slide accurately",
        "prompt_tone": "methodical, precise",
        "restrictions": [
            "Do not omit any necessary detail for replicating the slide accurately.",
            "Always include concrete guidance for content, structure, and phrasing.",
            "Do not leave prompts ambiguous or open-ended.",
            "English only."
        ]
\end{verbatim}
}

\subsection{Persona B: Balanced Editor}
{\small
\begin{verbatim}
        "prompt_verbosity": "medium",
        "prompt_detail_level": "clear, actionable prompt at slide or bullet level; may include some suggestions for 
        phrasing but no full tables or exact numbers in the prompt itself; maximum 4 sentences in the prompt",
        "prompt_tone": "efficient, professional",
        "restrictions": [
            "Do not reference files the downstream assistant cannot access.",
            "Avoid lengthy paragraphs; keep prompt focused on clarity and actionable guidance.",
            "English only."
        ]
\end{verbatim}
}

{\small
\subsection{Persona C: Executive}
\begin{verbatim}
        "prompt_verbosity": "short and concise",
        "prompt_detail_level": "generate a very succinct prompt; only one sentence of prompt.",
        "prompt_tone": "brief, strategic, professional",
        "restrictions": [
            "Do not reference any external files or resources the downstream assistant cannot access.",
            "Do not omit any necessary information for replicating the slide accurately.",
            "Do not give step-by-step instructions; only describe the conceptual or structural change.",
            "English only."
        ]
\end{verbatim}
}

\section{Examples of User Simulated Editing Prompts}

The following are the first turn prompts for Paper 179. 

\subsection{Kimi~K2 Examples}

\subsubsection{Executive} 
Add a dedicated slide titled 'Future Direction' that lists the three envisioned research directions—high-level vision tokenizer, asymmetric generative model, and low-level vision tasks—exactly as presented in the gold deck.

\subsubsection{Balanced} 
Replace the current 'Problem Statement' slide with a concise 'Motivation' slide that follows the gold deck's flow: open with the promise of INRs (arbitrary resolution, memory efficiency, detail capture), then highlight the two-type split (MLP-based slow vs. feature-grid memory-hungry) and conclude that both are unfriendly to low-end devices. Keep bullets short and avoid tables or deep metrics. 

\subsubsection{Granular} 
Add a new slide titled 'Limitations' after the Conclusion slide. The slide should contain the following content: a centered title 'Limitations', followed by three bullet points: (1) 'Gaussian initialization still needs 1-2 auxiliary MLP layers, preventing pure-Gaussian deployment.' (2) 'RVQ codebooks add memory overhead at low bitrates; ~0.5 MB per codebook.' (3) 'Trained on single images; no extension yet to video or 3D.' Ensure each bullet is concise and matches the exact phrasing provided.

\subsection{GPT Examples}

\subsubsection{Executive} 
Revise the existing slides by adding a new early slide titled “Background: Implicit Neural Representations” that briefly explains INR as a coordinate-to-signal neural function with examples of signals (image/video/3D object), lists key advantages (arbitrary resolution, memory efficiency, detail retention for tasks like inpainting/deblurring/denoising), and cites SIREN and NeRV, matching the level of detail and structure of the background description while keeping style consistent with the rest of the deck.

\subsubsection{Balanced} 
Revise the early-background portion of the deck to better align with the structure and emphasis of the reference slides on implicit neural representations and Gaussian splatting. Specifically, split the current combined “Problem Statement / Related Work Context” into (1) a short background slide introducing INRs as continuous coordinate-to-RGB functions and listing their general advantages (arbitrary resolution, memory efficiency, detail preservation, and example applications), and (2) a separate slide classifying image INRs into MLP-based and feature-grid-based methods with their respective pros/cons for training speed and GPU memory. Then add or adapt a slide titled “Motivation: Gaussian Splatting” that briefly introduces 3D Gaussian Splatting as an explicit 3D representation with differentiable rasterization, high visual quality, and real-time rendering, and clearly states that the goal is to bring these properties (efficient training, fast decoding, memory friendliness) to single-image representation. Keep bullets concise and conceptual, and avoid tables or detailed numbers on these background slides. 

\subsubsection{Granular} 
Revise and expand the introductory slides of the previous deck so that they accurately reflect the structure, content, and emphasis of the source presentation, while keeping the existing slide order and later technical slides intact.

Make the following concrete changes:

1. **Replace the current “Problem Statement” slide with a slide titled exactly:**
   
   `Background: Implicit Neural Representations`
   
   Use this structure and wording:
   - Title: `Background: Implicit Neural Representations`
   - Bullet block 1 (what INRs are):
     - Top-level bullet: `Parametrize a signal as a continuous function`
       - Sub-bullet: `Input: coordinate`
       - Sub-bullet: `Function: neural network`
       - Sub-bullet: `Output: RGB values, density`
   - Bullet block 2 (advantages):
     - Top-level bullet: `Advantages:`
       - Sub-bullet: `Arbitrary resolution → signal super-resolution`
       - Sub-bullet: `Memory efficient → signal compression`
       - Sub-bullet: `Capture, retain and infer signal details → signal inpainting, deblurring, denoising, …`
   - At the bottom of the slide, add a one-line references block:
     - `[1] Vincent Sitzmann et al. Implicit Neural Representations with Periodic Activation Functions, NeurIPS 2020.`
     - `[2] Hao Chen et al. NeRV: Neural Representations for Videos, NeurIPS 2021.`
   - Remove the current bullets about “INRs achieve high visual quality (10–1000 FPS) but require substantial GPU memory” and “Current INR methods (WIRE, INGP) suffer from…” from this slide. Those points will be moved to a separate follow-up slide in the next step.
   - Do not add images here; the slide text should be sufficient to replicate.

2. **Insert a new slide immediately after the revised INR background slide, titled:**
   
   `Background: Implicit Neural Representations (Image INRs)`
   
   This slide should explicitly cover the two INR types and the limitations for low-end devices, aligning with the gold deck and the paper’s Section 2.1.
   - Title: `Background: Implicit Neural Representations (Image INRs)`
   - Introductory sentence as a top-level bullet: `Two types in image INRs:`
     - Sub-bullet group 1: `MLP-based INRs:`
       - Sub-bullet: `Take position-encoded spatial coordinates as input of an MLP to learn RGB values`
       - Sub-bullet: `Long training times`
       - Sub-bullet: `Slow decoding speed`
       - Sub-bullet: `High GPU memory consumption`
     - Sub-bullet group 2: `Feature grid-based INRs:`
       - Sub-bullet: `Adopt multi-resolution feature grids (e.g., hash tables, quadtrees) plus a compact MLP`
       - Sub-bullet: `Fast training and inference`
       - Sub-bullet: `Higher GPU memory consumption`
   - Concluding bullet at the bottom of the list: `Low-end devices with limited memory: unfriendly!`
   - Add a reference line at the bottom:
     - `[3] Thomas Müller et al. Instant Neural Graphics Primitives with a Multiresolution Hash Encoding, SIGGRAPH 2022.`

3. **Revise the existing “GaussianImage Solution” slide into a “Motivation: Gaussian Splatting” slide that mirrors the gold deck’s motivation content, and then move the current solution bullets to a separate “GaussianImage Overview” slide.**
   
   a) Change the title `GaussianImage Solution` to:
   - `Motivation: Gaussian Splatting`
   
   b) Replace the current bullet points on this slide with the following bullets that justify why Gaussian Splatting is considered:
   - Top-level bullet: `The characteristics of advanced neural image representation:`
     - Sub-bullet: `Efficient training`
     - Sub-bullet: `Fast decoding`
     - Sub-bullet: `Friendly GPU memory usage`
   - Top-level bullet: `Gaussian Splatting in 3D scene reconstruction:`
     - Sub-bullet: `Explicit 3D Gaussian representations and differentiable tile-based rasterization`
     - Sub-bullet: `High visual quality with competitive training times`
     - Sub-bullet: `Real-time rendering capabilities`
   - Add reference line at the bottom:
     - `[4] Bernhard Kerbl et al. 3D Gaussian Splatting for Real-Time Radiance Field Rendering, ACM Transactions on Graphics 2023.`
   - Remove the performance numbers and codec description from this slide; they will be relocated.

   c) Immediately after the “Motivation: Gaussian Splatting” slide, insert a new slide titled:
   - `GaussianImage Overview`
   
   On this slide, restate the method-level summary and the key result numbers (these currently live in the old “GaussianImage Solution” and “Conclusion” slides) in a compact, clearly structured way:
   - Introductory top-level bullet: `GaussianImage: a groundbreaking image representation paradigm based on 2D Gaussian Splatting`
   - Bullet group: `Key ideas:`
     - Sub-bullet: `2D Gaussian representation with 8 parameters per Gaussian (position, covariance, weighted color coefficients)`
     - Sub-bullet: `Accumulated blending-based rasterization (order-invariant, no depth sorting)`
     - Sub-bullet: `Vector-quantization-based image codec (RVQ on color, quantized covariance, FP16 positions)`
   - Bullet group: `Main benefits (compared with 3D GS and INR baselines):`
     - Sub-bullet: `$\approx$7.375× compression over 3D Gaussians (59 → 8 parameters per Gaussian)`
     - Sub-bullet: `At least 3× lower GPU memory usage than feature-grid INRs`
     - Sub-bullet: `$\approx$5× faster fitting time`
     - Sub-bullet: `1500–2000 FPS rendering and decoding speed, independent of parameter size`
     - Sub-bullet: `Rate–distortion performance comparable to COIN and COIN++, with better MS-SSIM and further gains using partial bits-back coding`

4. **Keep the existing later technical slides (“2D Gaussian Representation”, “Accumulated Blending”, “Compression Pipeline”, tables, ablation, and conclusion) unchanged in this edit, except that any formulas or text that already use `accumulated blending` or `accumulated summation` should remain consistent with that wording.**

The goal of this edit request is to bring the early part of the deck (background and motivation) into close alignment with the structure and phrasing in the gold presentation: first define INRs and their advantages, then discuss their image-specific variants and limitations, then introduce Gaussian Splatting as a motivating technique, and finally summarize GaussianImage at a high level before entering the detailed method slides. Ensure wording follows the bullet texts given above as closely as possible so that another assistant could recreate the slides verbatim from this request. 

\section{Qualitative Editing Examples}

\label{sec:baseline}
\begin{figure*}[ht]
  \vskip 0.2in
  \begin{center}
    \centerline{\includegraphics[width=\textwidth]{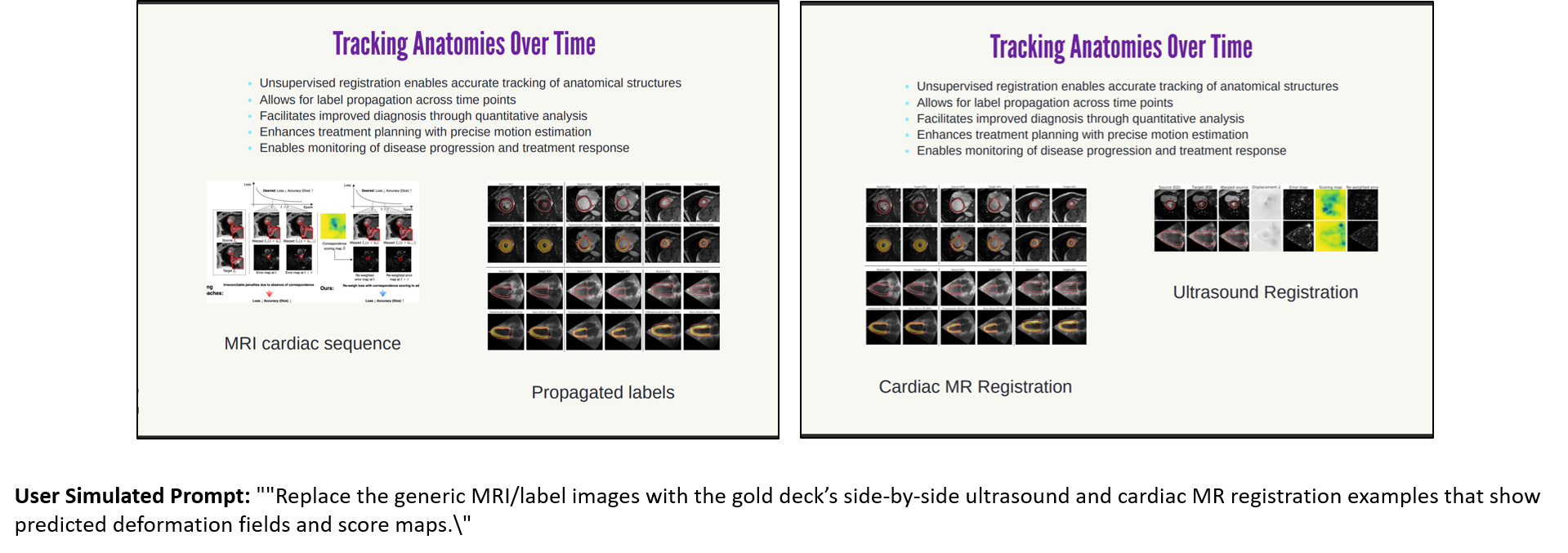}}
    \caption{
      Example of a slide before (left) and after (right) applying the user simulated prompt.}
    \Description{Example of a slide before (left) and after (right) applying the user simulated prompt.}
    \label{fig:example1}
  \end{center}
\end{figure*}

\begin{figure*}[ht]
  \vskip 0.2in
  \begin{center}
    \centerline{\includegraphics[width=\textwidth]{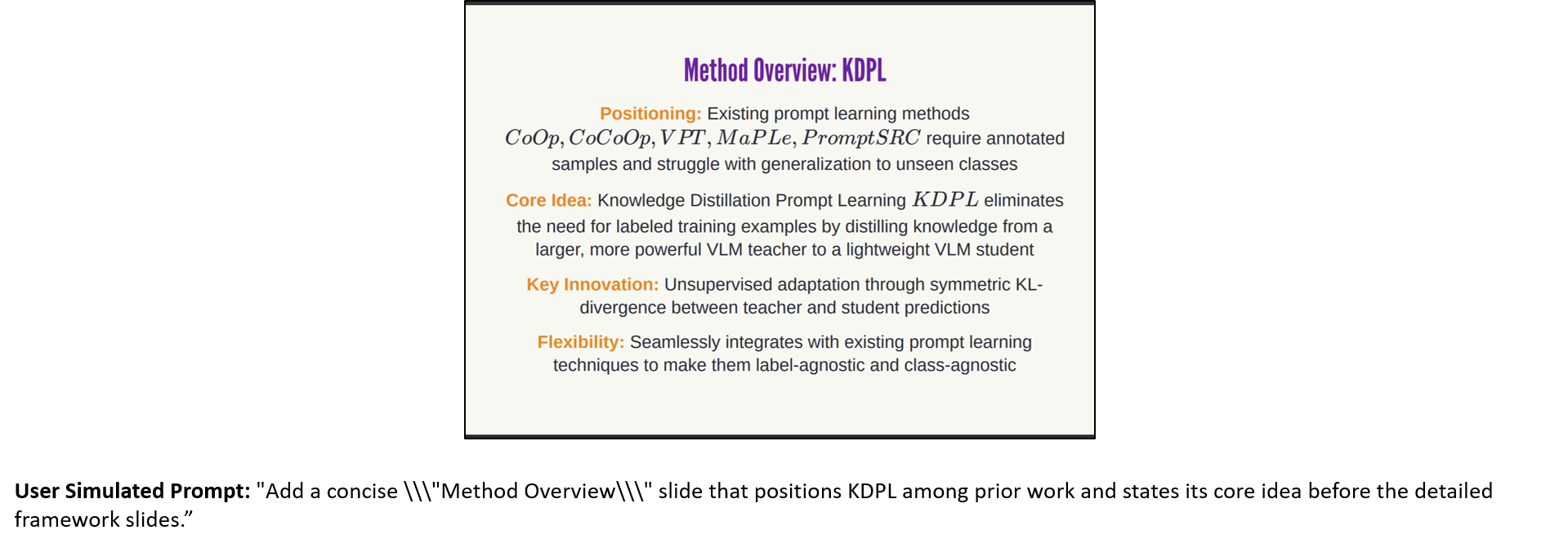}}
    \caption{
      Example of an added slide after applying the user simulated prompt.}
    \Description{Example of an added slide after applying the user simulated prompt.}
    \label{fig:example2}
  \end{center}
\end{figure*}

\end{document}